\documentclass[sigconf,natbib=false]{acmart}

\AtBeginDocument{%
  \providecommand\BibTeX{{%
    \normalfont B\kern-0.5em{\scshape i\kern-0.25em b}\kern-0.8em\TeX}}}

\copyrightyear{2020} 
\acmYear{2020} 
\setcopyright{rightsretained} 
\acmConference[KDD '20]{Proceedings of the 26th ACM SIGKDD Conference on Knowledge Discovery and Data Mining}{August 23--27, 2020}{Virtual Event, CA, USA}
\acmBooktitle{Proceedings of the 26th ACM SIGKDD Conference on Knowledge Discovery and Data Mining (KDD '20), August 23--27, 2020, Virtual Event, CA, USA}
\acmDOI{10.1145/3394486.3403296}
\acmISBN{978-1-4503-7998-4/20/08}

\usepackage[style=ACM-Reference-Format,maxbibnames=7,maxcitenames=1, backend=biber,uniquelist=false,natbib,doi=false,isbn=false,url=false,giveninits]{biblatex}
\renewbibmacro{in:}{}
\usepackage{subcaption}
\usepackage{pgf}
\usepackage{multirow}
\usepackage{siunitx}
\usepackage{algorithm, algorithmic}
\usepackage{booktabs}
\usepackage{enumitem}

\usepackage{amsmath,amsfonts,bm}

\def\eqref#1{equation~\ref{#1}}

\def\1{\bm{1}}

\def\eps{{\epsilon}}

\def\vd{{\bm{d}}}
\def\ve{{\bm{e}}}

\def\vr{{\bm{r}}}

\def\vx{{\bm{x}}}

\def\vz{{\bm{z}}}

\def\mA{{\bm{A}}}

\def\mD{{\bm{D}}}

\def\mH{{\bm{H}}}
\def\mI{{\bm{I}}}

\def\mQ{{\bm{Q}}}

\def\mZ{{\bm{Z}}}

\DeclareMathAlphabet{\mathsfit}{\encodingdefault}{\sfdefault}{m}{sl}
\SetMathAlphabet{\mathsfit}{bold}{\encodingdefault}{\sfdefault}{bx}{n}

\def\gN{{\mathcal{N}}}

\newcommand{\vpi}{\bm{\pi}}
\newcommand{\mPi}{\bm{\Pi}}
\newcommand{\PiApp}{\mPi^{(\eps)}}
\newcommand{\piApp}{\vpi^{(\eps)}}
\newcommand{\nnz}{\mathcal{N}^{k}}

\newcommand{\model}{PPRGo}

\usepackage{eqparbox}

\begin{document}

\title{Scaling Graph Neural Networks with Approximate PageRank}


\author{Aleksandar Bojchevski}
\authornote{Both authors contributed equally to this research.}
\email{a.bojchevski@in.tum.de}
\author{Johannes Gasteiger}
\authornotemark[1]
\email{j.gasteiger@in.tum.de}
\affiliation{%
  \institution{Technical University of Munich}
}

\author{Bryan Perozzi}
\email{bperozzi@acm.org}
\author{Amol Kapoor}
\author{Martin Blais}
\author{Benedek R\'ozemberczki}
\author{Michal Lukasik}
\affiliation{%
	\institution{Google Research}
}

\author{Stephan G{\"u}nnemann}
\email{guennemann@in.tum.de}
\affiliation{%
	\institution{Technical University of Munich}
}

\setlength{\abovedisplayskip}{1pt}
\setlength{\belowdisplayskip}{1pt}
\setlength{\abovedisplayshortskip}{1pt}
\setlength{\belowdisplayshortskip}{1pt}

\renewcommand{\shortauthors}{Bojchevski et al.}
\fancyhead{} 

\begin{abstract}
Graph neural networks (GNNs) have emerged as a powerful approach for solving many network mining tasks.
However, learning on large graphs remains a challenge -- many recently proposed scalable GNN approaches rely on an expensive message-passing procedure to propagate information through the graph.
We present the \model~model which utilizes an efficient approximation of information diffusion in GNNs resulting in significant speed gains while maintaining state-of-the-art prediction performance.
In addition to being faster, \model~ is inherently scalable, and can be trivially parallelized for large datasets like those found in industry settings.

We demonstrate that \model~ outperforms baselines in both distributed and single-machine training environments on a number of commonly used academic graphs.
To better analyze the scalability of large-scale graph learning methods, we introduce a novel benchmark graph with 12.4 million nodes, 173 million edges, and 2.8 million node features. We show that training \model~ from scratch and predicting labels for all nodes in this graph takes under 2 minutes on a single machine, far outpacing other baselines on the same graph.
We discuss the practical application of \model~ to solve large-scale node classification problems at Google.\footnote{You can find the code and data online: 
\url{https://www.daml.in.tum.de/pprgo}}
\end{abstract}

%
%



\maketitle

\section{Introduction}
\label{sec:introduction}
Graph Neural Networks (GNNs) excel on a wide variety of network mining tasks
from semi-supervised node classification and link prediction \cite{kipf2016semi, hamilton2017inductive, velickovic2017graph, zhang2018link}
to community detection and graph classification \cite{gilmer2017neural, niepert2016learning, chen2018supervised, al2019ddgk}.
The success of GNNs on academic datasets has generated significant interest in scaling these methods to larger graphs for use in real-world problems \cite{chen2018fastgcn,chen2018stochastic, hamilton2017inductive, gao2018large, ying2018graph,huang2018adaptive, sato2019constant,chiang19cluster}. 
Unfortunately, there are few large graph baseline datasets available; apart from a handful of exceptions \cite{ying2018graph,chiang19cluster}, the scalability of most GNN methods has been demonstrated on graphs with fewer than 250K nodes. Moreover, the majority of existing work focuses on improving scalability on a single machine. Many interesting network mining problems involve graphs with billions of nodes and edges that require distributed computation across many machines. As a result, we believe most of the current literature does not accurately reflect the major challenges of large scale GNN computing.

The main scalability bottleneck of most GNNs stems from the recursive message-passing procedure that propagates information through the graph. Computing the hidden representation for a given node requires joining information from its neighbors, and the neighbors in turn have to consider \textit{their own} neighbors, and so on. This process leads to an expensive neighborhood expansion, growing exponentially with each additional layer.

In many proposed GNN pipelines, the exponential growth of neighborhood size corresponds to an exponential IO overhead. A common strategy for scaling GNNs is to sample the graph structure during training, e.g. sample a fixed number of nodes from the $k$-hop neighborhood of a given node to generate its prediction \cite{hamilton2017inductive, ying2018graph}. The key differences between many scalable techniques lies in the design of the sampling scheme. For example, \citet{chen2018fastgcn} directly sample the receptive field for each layer using importance sampling, while \citet{chen2018stochastic} use the historical activations of the nodes as a control variate. \citet{huang2018adaptive} propose an adaptive sampling strategy with a trainable sampler per layer, and \citet{chiang19cluster} sample a block of nodes corresponding to a dense subgraph identified by the clustering algorithm METIS \cite{karypis1998fast}. Because these approaches still rely on a multi-hop message passing procedure, there is an extremely steep trade-off between runtime and accuracy. Unfortunately, for many of the proposed methods sampling does not directly reduce the number of nodes that need to be retrieved, since e.g. we have first have to compute the importance scores \cite{chen2018fastgcn}.

Recent work shows that personalized PageRank \cite{jeh2003ScalingPW} can be used to directly incorporate multi-hop neighborhood information of a node without explicit message-passing \cite{gasteiger2018combining}.
Intuitively, propagation based on personalized PageRank corresponds to infinitely many neighborhood aggregation layers where the node influence decays exponentially with each layer.
However, as proposed, \citet{gasteiger2018combining}'s approach does not easily scale to large graphs since it performs an expensive variant of power iteration during training. 

In this work, we present \model, a GNN model that scales to large graphs in both single and multi-machine (distributed) environments by using an adapted propagation scheme based on \emph{approximate} personalized PageRank.
Our approach removes the need for performing expensive power iteration during each training step by utilizing the (strong) localization properties \cite{gleich2015localization, nassar2015strong} of personalized PageRank vectors for real-world graphs. These vectors can be readily approximated with sparse vectors and efficiently pre-computed in a distributed manner \cite{andersen2006local}.
Using the sparse pre-computed approximations we can maintain the influence of relevant nodes located multiple hops away without prohibitive message-passing or power iteration costs.
We make the following contributions:
\begin{itemize}[leftmargin=0.25cm,itemindent=.25cm,labelwidth=\itemindent,labelsep=0cm,align=left,topsep=0pt,itemsep=-1ex,partopsep=1ex,parsep=1ex]
    \item We introduce the \model~ model based on approximate personalized PageRank. On a graph of over 12 million nodes, \model~ runs in under 2 minutes on a single machine, including pre-processing, training and inference time.
    \item We show that \model~ scales better than message-passing GNNs, especially with distributed training in a real-world setting.
    \item 
    We introduce the \emph{MAG-Scholar} dataset (12.4M nodes, 173M edges, 2.8M node features), a version of the Microsoft Academic Graph that we augment with "ground-truth" node labels. The dataset is orders of magnitude larger than many commonly used benchmark graphs.
    \item Most previous work exclusively focuses on training time. We also show a significantly reduced \textit{inference} time and furthermore propose sparse inference to achieve an additional 2x speed-up.
\end{itemize}
\section{Background}
\subsection{GNNs and Message-Passing}
\label{sec:ppnp}
Many proposed GNN models can be analyzed using the message-passing framework proposed by \citet{gilmer2017neural} or other similar frameworks \cite{battaglia2018relational, wu2019comprehensive, chami2020machine}. Typically, the computation is carried out in two phases: (i) messages are propagated along the neighbors; and (ii) the messages are aggregated to obtain the updated representations. 
At each layer, transformation of the input (e.g. linear projection plus a non-linearity) is coupled with aggregation/propagation among the neighbors (e.g. averaging).
Increasing the number of layers is desirable since: (i) it allows the model to incorporate information from more distant neighbors; and (ii)
it enables hierarchical feature extraction and thus the learning of richer node representations.

However, this has both computational and modelling consequences.
First, the recursive neighborhood expansion at each layer implies an exponential increase in the overall number of nodes we need to aggregate to produce the output at the final layer which is computationally prohibitive for large graphs.\footnote{For large graphs on distributed storage, just gathering the required neighborhood data requires many expensive remote procedure calls that greatly increase run time.}
Second, it has been shown \cite{li2018deeper, xu2018representation} that naively stacking multiple layers may suffer from over-smoothing that can reduce predictive performance.

To tackle both of these challenges \citet{gasteiger2018combining} suggest decoupling the feature transformation from the propagation. In their PPNP model, predictions are first generated (e.g. with a neural network) for each node utilizing only that node's own features, and then propagated using an adaptation of personalized PageRank. Specifically, PPNP is defined as:
\begin{align}
	\label{eq:ppnp_full}
	\mZ = \textrm{softmax} \big( \mPi^\text{sym} \mH \big)
	,\quad\quad\quad
	\mH_{i,:} = f_\theta(\vx_i)
\end{align}
where $\mPi^\text{sym} = \alpha (\mI_n - (1-\alpha) \tilde{\mA})^{-1}$ is a symmetric propagation matrix, $\tilde{\mA}=\mD^{-1/2}\mA\mD^{-1/2}$ is the normalized adjacency matrix with added self-loops, $\alpha$ is a teleport (restart) probability, $\mH$ is a matrix where each row is a vector representation for a specific node, and $\mZ$ is a matrix where each row is a prediction vector for each node, after propagation. The local per-node representations $\mH_{i,:}$ are generated by a neural network $f_\theta$ that processes the features $\vx_i$ of every node $i$ independently.
The responsibility for learning good representations is delegated to $f_\theta$, while $\mPi^\text{sym}$ ensures that the representations are smoothly changing w.r.t. the graph.

Because directly calculating the dense propagation matrix $\mPi^\text{sym}$ in \autoref{eq:ppnp_full} is inefficient, the authors propose a variant of power iteration to compute the final predictions instead. Unfortunately, even a moderate number of power iteration evaluations (e.g. \citet{gasteiger2018combining} used $K=10$ to achieve a good approximation) is prohibitively expensive for large graphs since they need to be computed during each gradient-update step. Moreover, despite the fact that $\tilde{\mA}$ is sparse, graphs beyond a certain size cannot be stored in memory.

\subsection{Personalized PageRank and Localization}
\label{sec:ppr_localization}

Since it is more amenable to efficient approximation we analyze the personalized PageRank matrix $\mPi^\text{ppr} = \alpha(\mI_n - (1-\alpha) \mD^{-1}\mA)^{-1}$. Each row $\vpi(i):=\mPi^\text{ppr}_{i, :}$ is equal to the personalized (seeded) PageRank vector of node $i$.
PageRank and its many variants \cite{1998ThePC, jeh2003ScalingPW, wang2005DirichletP} have been extensively studied in the literature. Here we are interested in efficient and scalable algorithms for computing (an approximation) of personalized PageRank. Luckily, given the broad applicability of PageRank, many such algorithms have been developed \cite{fogaras2004TowardsSF, andersen2006local, andersen2008LocalCO, gleich2015localization,lofgren2016PersonalizedPE,wang2016HubPPREI,wang2017FORASA, wei2018TopPPRTP, fujiwara2013FastAE}.

Random walk sampling \cite{fogaras2004TowardsSF} is one such approximation technique. While simple to implement, in order to guarantee at most $\eps$ absolute error with probability of $1-1/n$ we need $O(\frac{\log n}{\eps^2})$ random walks.
Forward search \cite{andersen2006local, gleich2015localization} and backward search \cite{andersen2008LocalCO} can be viewed as deterministic variants of the random walk sampling method. Given a starting configuration, the PageRank scores are updated by traversing the out-links (respect., in-links) of the nodes. 

For this work we adapt the approach by \citet{andersen2006local} since it offers a good balance of scalability, approximation guarantees, and ease of distributed implementation.
They show that $\vpi(i)$ can be weakly approximated with a low number of non-zero entries using a scalable algorithm that applies a series of push operations which can be executed in a distributed manner.

When the graph is strongly connected $\vpi(i)$ is non-zero for all nodes. Nevertheless, we can obtain a good approximation by truncating small elements to zero since most of the probability mass in the personalized PageRank vectors $\vpi(i)$ is \emph{localized} on a small number of nodes \cite{nassar2015strong,gleich2015localization,andersen2006local}. Thus, we can approximate $\vpi(i)$ with a sparse vector and in turn approximate $\mPi^\text{ppr}$ with a sparse matrix.

Once we obtain an approximation $\PiApp$ of $\mPi^{\text{ppr}}$ we can either use it directly to propagate information, or we can renormalize it via $\mD^{1/2} \PiApp \mD^{-1/2}$ to obtain an approximation of the matrix $\mPi^\text{sym}$.
 
\subsection{Related work}
\label{sec:related}
\textbf{Scalability}. GNNs were first proposed in \citet{gori2005new} and in \citet{scarselli2009graph} and have since emerged as a powerful approach for solving many network mining tasks \cite{bruna2013spectral,defferrard2016convolutional,kipf2016semi, gilmer2017neural, velickovic2017graph, scarselli2009graph, abu2018n, pmlr-v97-abu-el-haija19a}.
Most GNNs do not scale to large graphs since they typically need to perform a recursive neighborhood expansion to compute the hidden representations of a given node. While several approaches have been proposed to improve the efficiency of graph neural networks \cite{chen2018fastgcn,chen2018stochastic, hamilton2017inductive, gao2018large, ying2018graph,huang2018adaptive, sato2019constant, wu2019simplifying,chiang19cluster}, the scalability of GNNs to massive (web-scale) graphs is still under-studied.
As we discussed in \autoref{sec:introduction} the most prevalent approach to scalability is to sample a subset of the graph, e.g. based on different importance scores for the nodes \cite{hamilton2017inductive, ying2018graph, gao2018large, sato2019constant, chiang19cluster}.\footnote{
The importance sampling score by \citet{ying2018graph} can be seen as an approximation of the non-personalized PageRank, however the number of random walks required to achieve a good approximation is relatively high \cite{fogaras2004TowardsSF} making it a suboptimal choice.}
Beyond sampling, \citet{gao2018large} collect the representations from a node's neighborhood into a matrix, sort independently along each column/feature, and use the $k$ largest entries as input to a $1$-dimensional CNN.
These techniques all focus on single-machine environments with limited (GPU) memory.

\citet{buchnik2018bootstrapped} propose feature propagation which can be viewed as a simplified linearized GNN. They perform graph-based smoothing as a preprocessing step (before learning) to obtain diffused node features which are then used to train a logistic regression classifier to predict the node labels. 
\citet{wu2019simplifying} propose an equivalent simple graph convolution (SGC) model 
and diffuse the features by multiplication with the k-th power of the normalized adjacency matrix.
However, node features are often high dimensional, which can make the preprocessing step computationally expensive. More importantly, while node features are typically sparse, the obtained diffused features become denser, which significantly reduces the efficiency of the subsequent learning step. Both of these approaches are a special case of the PPNP model \cite{gasteiger2018combining} which experimentally shows higher classification performance \cite{gasteiger2018combining, fey2019fast}.

\textbf{Approximating PageRank}. Recent approaches combine basic techniques to create algorithms with enhanced guarantees \cite{lofgren2016PersonalizedPE,wang2016HubPPREI,wang2017FORASA}. For example \citet{wei2018TopPPRTP} propose the TopPPR algorithm combining the strengths of random walks and forward/backward search simultaneously. They can compute the top $k$ entries of a personalized PageRank vector up to a given precision using a filter-and-refine paradigm. Another family of approaches \cite{fujiwara2013FastAE} are based on the idea of maintaining upper and lower bounds on the PageRank scores which are then used for early termination with certain guarantees. For our purpose the basic techniques are sufficient.

\section{The \model~Model}
\label{sec:pprgo_model}
\begin{figure}[t!]
	\centering
	\includegraphics[width=0.85\linewidth]{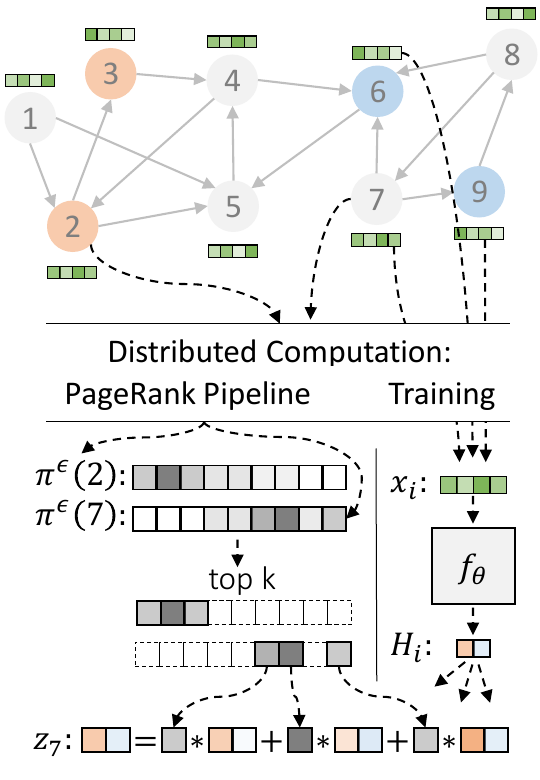}
	\caption{
		An illustration of \model. For each node $i$ we pre-compute an approximation of its personalized PageRank vector $\piApp(i)$.
		The approximation is computed efficiently and in parallel using a distributed batch data processing pipeline.
		The final prediction $\vz_i$ is then generated as a weighted average of the local (per-node) representations $\mH_{j, :} = f_\theta(\vx_j)$ for the top $k$ nodes ordered by largest personalized PageRank score $\vpi(i)_j$.
		To train the model $f_\theta(\cdot)$ that maps node attributes $\vx_i$ to local representations $\mH_i$, we only need the personalized PageRank vectors of the training nodes and attributes of the their respective top $k$ neighbors. The model is trained in a distributed manner on multiple batches of data in parallel.
	}
	\label{fig:our_model}
\end{figure}

The design of our model is motivated by: (i) the insights from \autoref{sec:ppnp}, namely that we can decouple the feature transformation from the information propagation, and (ii) the insights from \autoref{sec:ppr_localization}, namely that we can approximate $\mPi^\text{ppr}$ with a sparse matrix.
Analogous to \autoref{eq:ppnp_full} we define the final predictions of our model (see \autoref{fig:our_model}):
\begin{align}
\label{eq:our_model_all}
\mZ = \textrm{softmax} \big( \PiApp \mH \big)
,\quad\quad\quad
\mH_{i,:} = f_\theta(\vx_i)
\end{align}
where $\PiApp$ is a sparse approximation of $\mPi^\text{ppr}$. To obtain each row of $\PiApp$ we adapt the push-flow algorithm described in \citet{andersen2006local}.
We additionally truncate $\PiApp$ to contain only the top $k$ largest entries for each row. That is, for each node $i$ we only consider the set of nodes with top $k$ largest scores according to $\vpi(i)$.
Combined, the predictions for a given node $i$ are:
\begin{align}
\label{eq:our_model_single_node}
\vz_i = 
\textrm{softmax} \bigg( \sum_{j \in \nnz(i)} \piApp(i)_j \mH_j \bigg)
\end{align}
where $\nnz(i)$ enumerates the indices of the top $k$ largest non-zero entries in $\piApp(i)$. \autoref{eq:our_model_single_node} highlights that we only have to consider a small number of other nodes to compute the final prediction for a given node. Furthermore, this definition allows us to explicitly trade-off scalability and performance by increasing/decreasing the number of neighbors $k$ we take into account. We can achieve a similar trade-off by changing the threshold $\eps$ which effectively controls the norm of the residual. We show the pseudo-code for computing $\piApp$ in Algorithm \ref{alg:invPPR}. For further details see \autoref{sec:implementation}.

\begin{algorithm}[!h]
    \caption{Approximate personalized PageRank $(G, \alpha, t, \epsilon)$ \cite{andersen2006local}}
    \label{alg:invPPR}
    
    \begin{algorithmic}[1]
    \REQUIRE Graph $G$, teleport prob. $\alpha$, target node $t$, max. residual $\epsilon$
    \STATE Initialize the (sparse) estimate-vector $\piApp = \bm 0$ and the (sparse) residual-vector $\vr = \alpha \cdot \ve_t$ (i.e.~$\ve_t = 1$, $\ve_v=0, v\neq t$)
    \WHILE[$\vd_v$ is the out-degree]{$\exists v \,s.t.\, \vr_v> \alpha \cdot \epsilon \cdot \vd_v$}
    \STATE      $\piApp_v \mathrel{+}= \vr_v$
    \STATE      $\vr_v=0$
    \STATE $ m = (1 - \alpha) \cdot \vr_v / \vd_v$
    \FOR[$v$'s outgoing neighbors]{$u\in \gN_G^{\textrm{out}}(v)$}
        \STATE $\vr_u \mathrel{+}= m $
    \ENDFOR
    \ENDWHILE
    \RETURN $\piApp$
    \end{algorithmic}
\end{algorithm}    

In contrast to the PPNP model, a big advantage of \model~ is that we can pre-compute the sparse matrix $\PiApp$ before we start training. Pre-computation allows \model~ to calculate the training and inference predictions in $O(k)$ time, where $k \ll N$, and $N$ is number of nodes. Better still, for training we only require the rows of $\PiApp$ corresponding to the training nodes and the representations $f_\theta(\vx_i)$ of their top-k neighbors.
Furthermore, our model lends itself nicely to batched computation. For example, for a batch of nodes of size $b$ we have to load in memory the features of at most $b\cdot k$ nodes. In practice, this number is smaller than $b\cdot k$ since the nodes that appear in $\nnz(i)$ often overlap for the different nodes in the batch. We discuss the applicability and limitations of \model~ in \autoref{sec:limitations}. 

\subsection{Effective Neighborhood, $\alpha$ and $k$}
From the definition of personalized PageRank we can conclude that the hyper-parameter $\alpha$ controls the amount of information we are incorporating from the neighborhood of a node.
Namely, for values of $\alpha$ close to $1$ the random walks return (teleport) to the node $i$ more often and we are therefore placing more importance on the immediate neighborhood of the node. As the value of $\alpha$ decreases to $0$ we instead give more and more importance to the extended (multi-hop) neighborhood of the node. Intuitively, the importance of the $k$-hop neighborhood is proportional to $(1-\alpha)^k$. Note that the importance that each node assigns to itself (i.e. the value of $\vpi(i)_i$) is typically higher than the importance it assigns to the rest of the nodes. In conjunction with $\alpha$, we can modify the number of $k$ largest entries we consider to increase or decrease the size of the effective neighborhood. This stands in stark contrast to message-passing frameworks, where incorporating information from the extended neighborhood requires additional layers, thereby significantly increasing the computational complexity. 

\section{Scalability}
Here we discuss the properties of \model~ which make it suitable for large-scale classification problems occurring in industry.
\subsection{Node Classification in the Real World}
\label{sec:real_world}
The web is an incredibly rich data source and many different large graphs (potentially with \emph{hundreds of billions} of nodes and edges) can be derived from it.
Many web graphs have interesting node classification problems that can be addressed via semi-supervised learning. Their applications occur
across all media types and power many different Google products \cite{kannan2016smart,ssl_at_goog,Perozzi:2016:RGW:2939672.2939734}.
In web-scale datasets, the node sets are large, the graphs commonly have power-law degrees, 
the datasets change frequently, and labels can quickly become stale.
Therefore, having a model that trains as fast as possible is desirable to reduce the latency. Arguably even more important is having a model for which inference is as fast as possible, since inference is typically performed much more frequently than training in real-world settings. A low enough inference time may even open the door to using the model for online tasks, an impactful domain of problems where these models have limited penetration. Our proposed model, \model, ameliorates many of the difficulties associated with scaling these learning systems. We have successfully tested it on internal graphs with billions of nodes and edges.  

\subsection{Distributed Training}
In contrast to most previously proposed methods \cite{ying2018graph, hamilton2017inductive, wu2019simplifying} we utilize distributed computing techniques which significantly reduce the overall runtime of our method.
Our model is trained in two stages. First, we pre-compute the approximated personalized PageRank vectors using the distributed version of Algorithm \ref{alg:invPPR} (see \autoref{sec:implementation}). Second, we train the model parameters with stochastic gradient descent. Both stages are implemented in a distributed fashion. 

For the first stage we use an efficient batch data processing pipeline \cite{chambers2010flumejava} similar to MapReduce. Since we can compute the Page\-Rank vectors for every node in parallel our implementation easily scales to graphs with billions of nodes. 
Moreover, we can \textit{a priori} determine the number of iterations we need for achieving a desired approximation accuracy \cite{gleich2015localization, andersen2006local} which in turn means we can reliably estimate the runtime beforehand.

We implement \model~ in Tensorflow and optimize the parameters with \emph{asynchronous} distributed stochastic gradient descent. We store the model parameters on a parameter server (or several parameter servers depending on the model size) and multiple workers process the data in parallel. We use asynchronous training to avoid the communication overhead between many workers. Each worker fetches the most up-to-date parameters and computes the gradients for a mini-batch of data independently of the other workers. 

\subsection{Efficient Inference}
\label{sec:efficient_inference}
As discussed in \autoref{sec:pprgo_model} we only need to compute the approximate personalized PageRank vectors for the nodes in the training/validation set in order to train the model. In the semi-supervised classification setting these typically comprise only a small subset of all nodes (a few 100s or 1000s). However, during inference we still need to compute the PPR vector for every test node (see \autoref{eq:our_model_single_node}). Specifically, to predict the class label for $m < n$ test nodes we have to compute $\mZ = \textrm{softmax} \big( \mPi \mH \big)$ where $\mPi$ is a $m \times n$ matrix such that each row contains the personalized PageRank vector for a given test node, and $\mH$ is a $n \times c$ matrix of logits.
Even though the computation of each of these $m$ PPR vectors can be trivially parallelized, when $m$ is extremely large the overall runtime can still be considerable.
However, during inference we only use the PPR vectors a single time. In this case it is more efficient to circumvent this calculation and fall back to power iteration, i.e.
\begin{equation}
    \mQ^{(0)} = \mH, \qquad \qquad
    \mQ^{(p+1)}  = (1 - \alpha)  \mD^{-1} \mA \mQ^{(p)} + \alpha \mH.
\label{eq:power_iter}
\end{equation}
We furthermore found that, as opposed to training, during inference only very few (i.e. 1-3) steps of power iteration are necessary until accuracy improvements level off (see \autoref{sec:exp_inf}). Hence we only need very few sparse matrix-matrix multiplications for inference, which can be implemented very efficiently.

Since this truncated power iteration is very fast to compute, the neural network $f_\theta$ quickly becomes the limiting factor for inference time, especially if it is computationally expensive (e.g. a deep ResNet architecture \citep{he_deep_2016} or recurrent neural network (RNN)). With \model, we can leverage the graph's homophily to reduce the number of nodes that need to be analyzed. Since nearby nodes are likely to be similar we only need to calculate predictions $\mH$ for a small, randomly chosen fraction of nodes. Setting the remaining entries to zero we can smooth out these sparse labels over the rest via \autoref{eq:power_iter}.

In the very sparse case, using homophily to limit the number of needed predictions can be viewed as a label propagation problem with labels given by logits $\mH$. In the context of label propagation, the power iteration in \autoref{eq:power_iter} is a common algorithm known as "label propagation with return probability". This algorithm is known to perform well; we find that we can almost match the performance of full prediction with only a small fraction (e.g. \SI{10}{\percent} or \SI{1}{\percent}) of logits (see \autoref{sec:exp_inf}). Overall, this approach allows us to reduce the runtime even below a model that ignores the graph and instead considers each node independently, without sacrificing accuracy.
\section{Experiments}
\textbf{Setup.} We focus on semi-supervised node classification on attributed graphs
and demonstrate the strengths and scalability of \model~ in both distributed and single-machine environments.
To best align with real use cases we only use 20 $\cdot$ number of classes uniformly sampled (non-stratified) training nodes.
We fix the value of the teleport parameter to a common $\alpha=0.25$ for all experiments except the unusually dense Reddit dataset, where $\alpha = 0.5$. For details regarding training, hyperparameters, and metrics see \autoref{sec:exp_details} in the appendix.
We answer the following research questions:
\begin{itemize}[leftmargin=0.25cm,itemindent=.25cm,labelwidth=\itemindent,labelsep=0cm,align=left,topsep=0pt,itemsep=-1ex,partopsep=1ex,parsep=1ex]
    \item  What kind of trade-offs  between scalability and accuracy can we achieve with \model? (\autoref{sec:exp_scalability_v_performance})
    \item How effectively can we leverage distributed training? (\autoref{sec:exp_distributed_training})
    \item How much resources (memory, compute) does PPRGo need compared to other scalable GNNs? (\autoref{sec:exp_runtime_and_memory})
    \item How efficient is the proposed sparse inference scheme? (\autoref{sec:exp_efficient_inference})
\end{itemize}

\subsection{Large-Scale Datasets}
The majority of previous approaches are evaluated on a small set of publicly available benchmark datasets \cite{chen2018fastgcn,chen2018stochastic, hamilton2017inductive, gao2018large,huang2018adaptive, sato2019constant, wu2019simplifying, pmlr-v97-abu-el-haija19a}. The size of these datasets is relatively small, with the Reddit graph (233K nodes, 11.6M edges, 602 node features) \cite{hamilton2017inductive} typically being the largest graph used for evaluation.\footnote{
The Twitter geo-location datasets used in previous work \cite{wu2019simplifying} have limited usefulness for evaluating GNNs since
they have no meaningful graph structure, e.g. 70\% of the nodes in the Twitter-World dataset only have a self-loop and no other edges.
}
\citet{chiang19cluster} recently introduced the Amazon2M graph (2.5M nodes, 61M edges, 100 node features) which is large in terms of number of nodes, but tiny in terms of node feature size.\footnote{While larger benchmark graphs can be found in the literature, they either do not have node features or they do not have "ground-truth" node labels.}

\textbf{MAG-Scholar.}
To facilitate the development of scalable GNNs we create a new benchmark dataset based on the Microsoft Academic Graph (MAG) \cite{sinha2015AnOO}. Nodes represent papers, edges denote citations, and node features correspond to a bag-of-words representation of paper abstracts. We augmented the graph with "ground-truth" node labels corresponding to the papers' field of study. 

We extract the node labels semi-automatically by mapping the publishing venues (conferences and journals) to a field of study using metadata on the top venues from Google Scholar. We create two sets of labels for the same graph. Coarse-grained labels correspond to the following 8 coarse-grained fields of study: biology, engineering, humanities, medicine, physics, sociology, business, and other. We refer to this graph as MAG-Scholar-C. Fine-grained labels correspond to 253 fine-grained fields of study such as: architecture, epidemiology, geology, ethics, anthropology, linguistics, etc. The fine-grained labels make the classification problem more difficult. We refer to this graph as MAG-Scholar-F.

The resulting \emph{MAG-Scholar} graph is a few orders of magnitude larger then the commonly used benchmark graphs (12.4M nodes, 173M edges, 2.8M node features).
The graphs and the code to generate them will be made publicly available. See \autoref{sec:graph_construction} for a detailed description of the graph construction and node labelling process.

\subsection{Scalability vs. Accuracy Trade-off}
\label{sec:exp_scalability_v_performance}
The approximation parameter $\eps$ and the number of top-$k$ nodes are important hyper-parameters that modulate scalability and accuracy (see \autoref{eq:our_model_single_node}). We note that $\alpha$ and $k$ play similar roles, so we choose to analyze $k$ for a fixed $\alpha$.
To examine their effect on the performance of \model~we train our model on the MAG-Scholar-C graph for different values of $k$ and $\eps$. We repeat the experiment five times and report the mean performance.
We investigate two cases: a sparsely labeled scenario similar to industry settings (160 nodes), and an "academic" setting with many more labeled nodes (105415 nodes).

\begin{figure*}[t!]
    \begin{subfigure}[b]{0.49\linewidth}
	\centering
	\input{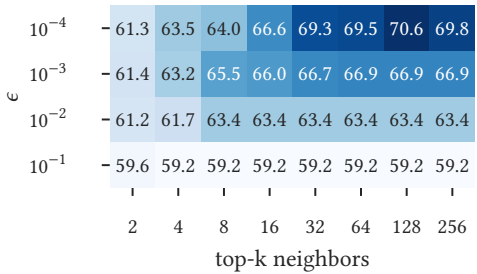}
	\caption{Sparsely labeled setting (160 nodes, \SI{0.0015}{\percent}) }
	\label{fig:k_vs_performance_magc_few_labels}
	\end{subfigure}
    \begin{subfigure}[b]{0.49\linewidth}
	\centering
	\input{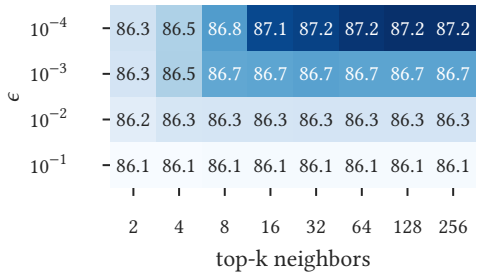}
	\caption{Setting with a large number of labeled nodes (105415 nodes, \SI{1}{\percent})}
	\label{fig:k_vs_performance_magc_many_labels}
	\end{subfigure}
	\caption{Mean accuracy (\%) over 5 runs on MAG-Scholar-C as we vary the number of neighbors 
	and the approx.\ parameter $\epsilon$.}
	\label{fig:k_vs_performance_magc}
\end{figure*}
\begin{figure}[h!]
    \centering
    \input{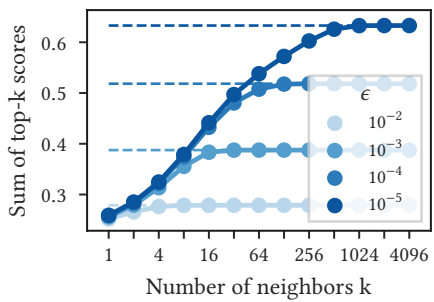}
	\caption{For each node in MAG-Scholar-C we calculate the sum of the top-$k$ largest scores in $\piApp(i)$ and we plot the average across all nodes for different values of $\epsilon$.
	The dashed line indicates $k=n$, i.e. the entire sum of $\piApp(i)$ averaged across nodes.
	The 95\% confidence intervals around the mean (estimated with bootstrapping) are too small to be visible.
	}
	\label{fig:topk_sum}
\end{figure}
As expected, we can see in \autoref{fig:k_vs_performance_magc}
that the performance consistently increases if we either use a more accurate approximation of the PageRank vectors (smaller $\eps$) or a larger number of top-k neighbors. This also shows that we can smoothly trade-off performance for scalability since models with higher value of $k$ and lower value of $\eps$ are computationally more expensive. For example, in the academic setting (\autoref{fig:k_vs_performance_magc_many_labels}) a model with $\eps=0.1, k=2$ had an overall (preprocessing + training + inference) runtime of 6 minutes,
while a model with $\eps=0.001, k=256$ had an overall runtime of
12 minutes.
Since many nodes are labeled (\SI{1}{\percent}) the difference between the highest accuracy (top right corner) and lowest accuracy (bottom left corner) is under \SI{2}{\percent} and the model is not sensitive to the hyperparameters.
In the sparsely labeled setting (\autoref{fig:k_vs_performance_magc_few_labels})
the choice of hyperparameters is more important and depends on the desired trade-off level
(slowest overall runtime was <2 minutes).

Interestingly, we can see on \autoref{fig:k_vs_performance_magc} that for any value of $\epsilon$ the performance starts to plateau at around top-$k=32$. 
The reason for this behavior becomes more clear by examining \autoref{fig:topk_sum}. Here, for each node $i$ we calculate the sum of the top-$k$ largest scores in $\piApp(i)$ and we plot the average across all nodes. We see that by looking at a very few nodes -- e.g. 32 out of 12.4 million -- we are able to capture the majority of the PageRank scores on average (recall that $\sum_j \piApp(i)_j \le 1$). Therefore, the curves in both \autoref{fig:k_vs_performance_magc} and \autoref{fig:topk_sum} plateau around the same value of $k$. These figures validate our approach of approximating the dense (but localized) personalized PageRank vectors with their respective sparse top-$k$ versions.
\begin{table*}[!t]
\caption{Breakdown of the runtime, memory, and predictive performance on a single machine for different models on the Reddit dataset. We use 820 (20 $\cdot$ \#classes) nodes for training.
We see that \model~ has a total runtime of less than 20\,s and is two orders of magnitude faster than SGC and Cluster-GCN. \model~ also requires less memory overall.}
\label{tab:acc_runtime_reddit}
\resizebox{\textwidth}{!}{
\begin{tabular}{@{}lc|cc|ccc|c|cc|c@{}}
\toprule
\multicolumn{1}{c}{\textbf{}} & \multicolumn{7}{c}{\textbf{Runtime (s)}}                                         & \multicolumn{2}{c}{\textbf{Memory (GB)}} & \textbf{Accuracy (\%)} \\ \midrule
\multicolumn{1}{c}{}          & Preprocessing & \multicolumn{2}{c|}{Training} & \multicolumn{3}{c|}{Inference}   & Total & RAM            & GPU            &          \\
                              &               & Per Epoch      & Overall     & Forward & Propagation & Overall &       &                &                &          \\ \midrule
Cluster-GCN          & \num{1175 +- 25}   & \num{4.77 +- 0.12}     & \num{953 +- 24}    & -                    & -                   & \num{186 +- 21}     & \num{2310 +- 40}     &\num{20.97 +- 0.15}   & \num{0.071 +- 0.006} & \num{17.1 +- 0.8} \\
SGC                  & \num{313 +- 9}     & \num{0.0026 +- 0.0002} & \num{0.53 +- 0.03} & -                    & -                   & \num{7470 +- 150}   & \num{7780 +- 150}    & \num{10.12 +- 0.03}  & \num{0.027}          & \num{12.1 +- 0.1} \\
PPRGo (1 PI step)    & \num{2.26 +- 0.04} & \num{0.0233 +- 0.0005}                 & \num{4.67 +- 0.10} & \num{0.341 +- 0.009}         & \num{5.85 +- 0.03} & \num{6.19 +- 0.04} & \num{13.1 +- 0.07}                     & \num{5.56 +- 0.019}                      & \num{0.073}          & \num{26.5 +- 1.9}      \\
PPRGo (2 PI steps)   & \num{2.22 +- 0.12} & \num{0.021 +- 0.003}                    & \num{4.1 +- 0.7}   & \num{0.43 +- 0.08}           & \num{10.1 +- 1.4}  & \num{10.5 +- 1.5}  & \num{16.8 +- 1.7}                      & \num{5.42 +- 0.18}                       & \num{0.073}          & \num{26.6 +- 1.8}      \\
\bottomrule
\end{tabular}
}
\end{table*}
\subsection{Distributed Training}
\label{sec:exp_distributed_training}
\begin{figure}[b!]
    \centering
    \input{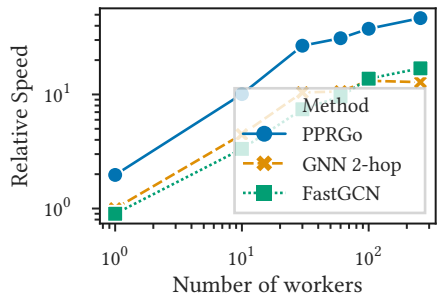}
	\caption{Relative speed in terms of number of gradient-update steps per second on the MAG-Scholar-F graph compared to the baseline method (GNN 2-hop, single worker). Both axes are on a log scale. \model~ is consistently the fastest method and can best utilize additional workers.}
	\label{fig:distributed_comparison_runtime}
\end{figure}
\begin{figure}[t!]
    \centering
    \input{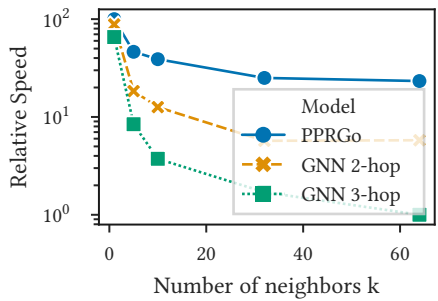}
	\caption{Relative speed comparison (num. gradient updates per second) between \model~ and multi-hop models for different values of $k$ on MAG-Scholar-F. Distributed training.}\vspace{-0.3em}
	\label{fig:k_vs_depth}
\end{figure}
In this section we aim to compare the performance of one-hop propagation using personalized PageRank and traditional multi-hop message passing propagation in a real distributed environment at Google. To make sure that the differences we observe are only due to the used model and not other factors, we implement simple 2-hop and 3-hop GNN models \cite{kipf2016semi}, which are also trained in a distributed manner using the same infrastructure as \model. 
Specifically, we make sure that both the multi-hop models and \model~ consider the same number of neighbors, e.g. if \model~ uses $k=64$ then the 2-hop model uses information from $8\times 8=64$ nodes from its first and second hop respectively. To select these neighborhoods we use a weighted sampling scheme similar to previous work \cite{ying2018graph}. Additionally, we implement a distributed version of FastGCN \cite{chen2018fastgcn} to evaluate the effect of different sampling schemes.

Our first observation is that there is no significant difference in terms of predictive performance between the different models (around 61\% accuracy). However, there is a significant difference in terms of runtime. On \autoref{fig:distributed_comparison_runtime} we show the speedup in terms of number of gradient-update steps per second on the MAG-Scholar-F graph as we increase the number of worker machines used for distributed training. Specifically, we show the relative speedup compared to the baseline method -- 2-hop GCN on a single worker. We see that \model~ is considerably faster than the baseline (note that both axes are on a log-scale). \model~ also requires fewer steps in total to converge. Moreover, the speedup gap between the 2 hop model and \model~ increases with the number of workers.
Crucially, since we have to fetch all neighbors to calculate their importance scores and since the runtime in the distributed setting is dominated by IO we see that FastGCN does not offer any significant scalability advantage over the baseline GNN 2-hop model.
In \autoref{sec:app_additional_experiments} we additionally analyze parallel efficiency, i.e. how well the different models utilize the additional workers. 

\model~ is able to process all top-$k$ neighbors at once, compared to the multi-hop models which have to recursively update the hidden representations. Therefore, while increasing the number of top-$k$ neighbors makes all model computationally more expensive, we expect the runtime of \model~ to increase the least. To validate this claim, we analyze the relative speed (number of gradient updates per second) compared to the slowest method at different values of $k$. The results in \autoref{fig:k_vs_depth} exactly match our intuition, and indeed the curve of \model~ has the smallest slope as we increase $k$, while the relative speed of the 2- and 3-hop GNNs deteriorate faster. FastGCN again matches GNN 2-hop, like it does in \autoref{fig:distributed_comparison_runtime} (not shown here).

\begin{table*}[t]
\caption{Single machine runtime (s), memory (GB), and accuracy (\%) for different models and datasets using 20 $\cdot$ \#classes training nodes. PPRGo shows comparable accuracy and scales much better to large datasets than its competitors.}
\label{tab:acc_runtime_datasets}
\resizebox{\textwidth}{!}{
\begin{tabular}{@{}lccc|ccc|ccc|ccc@{}}
\toprule
 & \multicolumn{3}{c}{\textbf{Cora-Full}} & \multicolumn{3}{c}{\textbf{PubMed}} & \multicolumn{3}{c}{\textbf{Reddit}} & \multicolumn{3}{c}{\textbf{MAG-Scholar-C}} \\
 & Time & Mem. & Acc. & Time & Mem. & Acc. & Time & Mem. & Acc. & Time & Mem. & Acc. \\  \midrule
Cluster-GCN                      & \num{84 +- 4}     & \num{2.435 +- 0.018} & \num{58.0 +- 0.7}                      & \num{54.3 +- 2.7} & \num{1.90 +- 0.03}   & \num{74.7 +- 3.0}                      & \num{2310 +- 50}    & \num{21.04 +- 0.15}  & \num{17.1 +- 0.8}      & >24h           & -           & -                 \\
SGC                              & \num{92 +- 3}     & \num{3.95 +- 0.03}   & \num{58.0 +- 0.8}                      & \num{5.3 +- 0.3}  & \num{2.172 +- 0.004} & \num{75.7 +- 2.3}                      & \num{7780 +- 140}   & \num{10.15 +- 0.03}  & \num{12.1 +- 0.1}      & >24h           & -           & -                 \\
APPNP                            & \num{10.7 +- 0.5} & \num{2.150 +- 0.019} & \num{62.8 +- 1.1} & \num{6.5 +- 0.4}  & \num{1.977 +- 0.004} & \num{76.9 +- 2.6} & -                   & OOM                  & - & -              & OOM         & -                 \\
PPRGo ($\epsilon=10^{-4}, k=32$) & \num{25 +- 3}     & \num{1.73 +- 0.03}   & \num{61.0 +- 0.7} & \num{3.8 +- 0.9}  & \num{1.626 +- 0.025} & \num{75.2 +- 3.3}               & \num{16.8 +- 1.7} & \num{5.49 +- 0.18}  & \num{26.6 +- 1.8}     & \num{98.6 +- 1.7} & \num{24.51 +- 0.04} & \num{69.3 +- 3.1} \\
PPRGo ($\epsilon=10^{-2}, k=32$) & \num{6.6 +- 0.5}  & \num{1.644 +- 0.013} & \num{58.1 +- 0.6} & \num{2.9 +- 0.5}  & \num{1.623 +- 0.017} & \num{73.7 +- 3.9}               & \num{16.3 +- 1.7} & \num{5.61 +- 0.06}  & \num{26.2 +- 1.8}     & \num{89 +- 5}     & \num{24.49 +- 0.05} & \num{63.4 +- 2.9} \\
\bottomrule
\end{tabular}
}
\end{table*}

\subsection{Runtime and Memory on a Single Machine}
\label{sec:exp_runtime_and_memory}
\textbf{Setup.} To highlight the benefits of \model~ we compare the runtime, memory, and predictive performance with SGC \cite{wu2019simplifying} and Cluster-GCN \cite{chiang19cluster}, two strong baselines that represent the current state-of-the-art scalable GNNs. Since SGC and Cluster-GCN report significant speedup over FastGCN \cite{chen2018fastgcn} and VRGCN \cite{chen2018stochastic} we omit these models from our comparison.

We run the experiments on Nvidia 1080Ti GPUs and on Intel CPUs (5 cores), using CUDA and TensorFlow. We run each experiment on five different random splits and report mean values and standard deviation. For SGC we use the second power of the graph Laplacian as suggested by the authors (i.e. we effectively have a 2-hop model). For \model~ we set $\epsilon=10^{-4}$ and $k=32$ following the discussion in \autoref{sec:exp_scalability_v_performance}.
We compute the overall runtime including the preprocessing time, the time to train the models, and the time to perform inference for all test nodes. This is in contrast to previous work which rarely report preprocessing time and almost never report inference time. For training, we report both the overall training time, as well as the time per epoch.

\noindent \textbf{Preprocessing time.} For each model, during preprocessing we perform the computation only for the training nodes. For SGC, the preprocessing step involves computing the diffused features using the second power of the graph Laplacian. We significantly optimized preprocessing for Cluster-GCN, resulting in node cluster computation with METIS \cite{karypis1998fast} becoming its main bottleneck. For \model, the preprocessing step involves computing the approximate personalized PageRank vectors using Algorithm \ref{alg:invPPR} and selecting the top $k$ neighbors. 

\noindent \textbf{Inference time.} For SGC, during inference we have to compute the diffused features for the test nodes (again using the second power of the graph Laplacian).
Following the implementation by the original authors of Cluster-GCN, we do not cluster the test nodes, but rather perform "standard" message-passing inference on the full graph.
For \model, as discussed in \autoref{sec:efficient_inference}, we run power iteration rather than computing the approximate PPR vectors for the test nodes. Two iteration steps were already sufficient to obtain good accuracy. We analyze the inference step in more detail in \autoref{sec:exp_efficient_inference}.

The results when training a model on the Reddit dataset (233K nodes, 11.6M edges, 602 node features) are summarized in \autoref{tab:acc_runtime_reddit}.
Both SGC and Cluster-GCN are several orders of magnitude slower than \model~. Interestingly, SGC is significantly slower w.r.t. inference time (since we have to compute the diffused features for all test nodes) while Cluster-GCN is significantly slower w.r.t. preprocessing and training time. The overall runtime of Cluster-GCN (\SI{2310}{\second}) and SGC (\SI{7470}{\second}) is in stark contrast to our proposed approach: under \SI{20}{\second}. Moreover, we see that the amount of memory used by \model~ is 4 times smaller compared to Cluster-GCN and 2 times smaller compared to SGC. Given that Cluster-GCN and SGC achieve significantly worse accuracy, the benefits of our proposed approach in terms of scalability are apparent.

We extend the above analysis to several other datasets. We chose two comparatively small academic graphs that are commonly used as benchmark datasets -- Cora-Full \cite{bojchevski2018deep} (18.7K nodes, 62.4K edges, 8.7K node features) and PubMed \cite{yang2015defining} (19.7K nodes, 44.3K edges, 0.5K node features) -- as well as our newly introduced MAG-Scholar-C dataset (10.5M nodes, 133M edges, 2.8M node features).
In addition to the two scalable baselines, we also evaluate how \model~ compares to the APPNP model \cite{gasteiger2018combining} which we build upon.
The results are summarized in \autoref{tab:acc_runtime_datasets}.
We can see that the performance of most models is comparable in terms of accuracy.
In most cases our proposed model \model~ has the smallest overall runtime and it always uses the least amount of memory. \model's comparatively long runtime on Cora-Full can be explained by its training set size: The training set is so large that PPRGo accesses more neighbors per batch than there are nodes in this graph, not leveraging the duplicate information. This can only happen with small graphs, for which runtime is not an issue.
We see that the APPNP model runs out of memory for even the moderately sized Reddit graph, highlighting the necessity of our approach.

More importantly, on the largest graph MAG-Scholar-C, we successfully trained \model~ from scratch and obtained the predictions for all test nodes in under 2 minutes, while Cluster-GCN and SGC were not able to finish in over 24 hours, with Cluster-GCN still stuck in preprocessing.

\subsection{Efficient Inference} \label{sec:exp_inf}
\label{sec:exp_efficient_inference}
\begin{figure}[t!]
    \centering
    \input{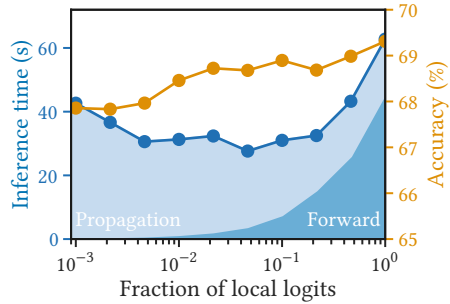}
    \caption{Accuracy and corresponding inference time (NN inference (dark blue) + propagation (light blue)) on MAG-Scholar-C w.r.t. the fraction of nodes for which local logits $\mH$ are inferred by the NN. PPRGo performs very well even if the NN is evaluated on very few nodes. We need more power iteration steps $p$ if we do fewer forward passes (see \autoref{fig:inf_acc}), increasing the propagation time. Note the logarithmic scale.}
    \label{fig:inf_opt_accrt}
\end{figure}

\begin{figure}[t!]
    \centering
    \input{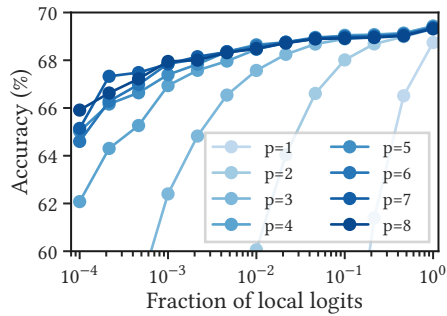}
    \caption{Accuracy on MAG-Scholar-C w.r.t. the fraction of nodes for which local logits $\mH$ are inferred and number of power iteration steps $p$. The fewer logits we calculate, the more power iteration steps we need for stabilizing the prediction. Note the logarithmic scale.}
    \label{fig:inf_acc}
\end{figure}

Inference time is crucial for real-world applications since a machine learning model needs to be trained only once, while inference is run continuously when the model is put into production. We found that PPRGo can achieve an accuracy of \SI{68.7}{\percent} with a \emph{single} power iteration step, i.e. without even calculating the PPR vectors. At this point, the neural network $f_\theta$ and not the propagation becomes the limiting factor. However, as described in \autoref{sec:efficient_inference}, we can reduce the neural network cost by only computing logits for a small, random subset of nodes. \autoref{fig:inf_opt_accrt} shows that the accuracy only reduces by around \num{0.6} percentage points when reducing the number of inferred nodes by a \emph{factor of 10}. We can therefore trade in a small amount of accuracy to significantly reduce inference time, in this case by \SI{50}{\percent}. With this approximation, PPRGo has a \emph{shorter} inference time than the forward pass of a simple neural network executed on each node independently. Furthermore, note that we use a rather simple feed-forward neural network in our experiments. This reduction will become even more dramatic in cases that leverage more computationally expensive neural networks for $f_\theta$. \autoref{fig:inf_acc} shows that when reducing the fraction of inferred nodes, the corresponding accuracy drops off earlier if we perform fewer power iteration steps $p$. Therefore, we need to increase the number of power iteration steps when we calculate fewer logits. This furthermore shows that subsampling logits would not be possible with methods that use locally sampled subgraphs (e.g. FastGCN). Note that we do not use this additional improvement in \autoref{tab:acc_runtime_reddit} and \ref{tab:acc_runtime_datasets}.

\section{Conclusion}
We propose a GNN for semi-supervised node classification that scales easily to graphs with millions of nodes. In comparison to previous work our model does not rely on expensive message-passing, making it well suited for use in large-scale distributed environments.
We can trade scalability and performance via a few intuitive hyperparameters.
To stimulate the development of scalable GNNs we present MAG-Scholar -- a new large-scale graph (12.4M nodes, 173M edges, and 2.8M node features) with coarse/fine-grained "ground-truth" node labels.  On this web-scale dataset \model~ achieves high performance in under 2 minutes (preprocessing + training + inference time) on a single machine. Beyond the single machine scenario, we demonstrate the scalability of \model~ in a distributed setting and show that it is more efficient compared to multi-hop models.

\section{Acknowledgements}
We would like to thank Chandan Yeshwanth for his assistance with conducting the experiments. This research was supported by the Deutsche Forschungsgemeinschaft (DFG) through the Emmy Noether grant GU 1409/2-1 and the TUM International Graduate School of Science and Engineering (IGSSE), GSC 81.

{
\tiny
\printbibliography
}
\clearpage
\appendix
\section{Appendix}
\subsection{Parallel Efficiency}
\label{sec:app_additional_experiments}
\begin{figure}[h!]
    \centering
    \input{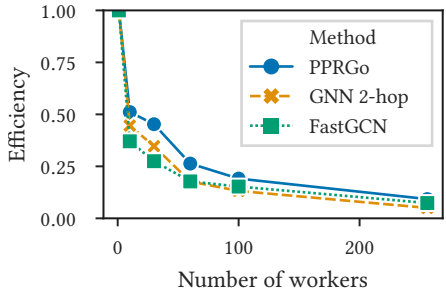}
	\caption{Parallel efficiency w.r.t. the number of distributed workers on MAG-Scholar-F for different models.}
	\label{fig:distributed_comparison_efficiency}
\end{figure}
To further investigate the performance of different models in the distributed training setting we also evaluate parallel efficiency. Intuitively, this efficiency measures how well we can utilize additional workers. Let $m_t$ be the number of steps per second using $t$ workers, then the parallel efficiency of a model is defined as $\frac{m_t}{m_1 \cdot t}$.
In \autoref{fig:distributed_comparison_efficiency} we see that \model~ achieves the best parallel efficiency.

\subsection{MAG-Scholar Graph Construction}
\label{sec:graph_construction}
First, we obtain the "raw" data from the Microsoft Academic Graph (MAG) \cite{sinha2015AnOO} repository, specifically we downloaded a snapshot of the data on 01.25.2019. We construct a graph where each node is a paper and the edges indicate citations/references between the papers. The node features are a bag-of-words representation of the paper's abstract. We preprocess the feature matrix by keeping only those words that appear in at least 5 abstracts. We preprocess the graph by keeping only the nodes that belong to the largest connected component. The resulting MAG-Scholar-F graph has 12.40393 million nodes, 2.78424 million features, and 173.050172 million edges. The MAG-Scholar-C graph has 10.54156 million nodes, 2.78424 million features, and 132.817644 million edges. To obtain the fields of study for each paper, we first create a mapping between a venue (i.e. conference or journal) and its respective field of study. Specifically, we consider the top-20 venues in each field of study according to Google Scholar\footnote{\url{https://scholar.google.com/citations?view_op=top_venues&hl=en}}. We manually match the same venues that have different tittles (e.g. because of abbreviations) in the MAG data compared to the Google Scholar data. These venues are categorized in 8 different coarse-grained categories (e.g. engeneering\footnote{\url{https://scholar.google.com/citations?view_op=top_venues&hl=en&vq=eng}}) and 253 different fine-grained categories (e.g. biophysics\footnote{\url{https://scholar.google.com/citations?view_op=top_venues&hl=en&vq=phy_biophysics}}) and we use them to define  coarse/fine-grained "ground-truth" labels for the nodes.

\subsection{Experimental Details}
\label{sec:exp_details}
We keep all \model~ hyperparameters constant across all datasets, except the value of the teleport parameter $\alpha=0.25$, which we set to $\alpha = 0.5$ for reddit. The feed-forward neural network has two layers, i.e. a single hidden layer of size 32. We use a dropout of 0.1 and set the weight decay to $10^{-4}$. We train for 200 epochs using a learning rate of $0.005$ and the Adam optimizer \cite{kingma2014adam} with a batch size of 512. To achieve a consistent setup across all models and datasets we always use the same number of epochs, use no early stopping and only evaluate validation accuracy after training. For the  validation set we randomly sample 10 times the number of training nodes.

We standardize the graphs as a preprocessing step, i.e. we choose only the subset of nodes belonging to the largest connected component and make the graph undirected and unweighted. We do not include dataset loading time in the overall runtime since it is the same for all models.

\subsection{Further Implementational Details}
\label{sec:implementation}
The pseudo code in Algorithm \ref{alg:invPPR} shows how we compute the approximate personalized PageRank based on \cite{andersen2006local}. For single-machine experiments we implement the algorithm as described in Python using Numba for acceleration (not parallelized). In the distributed setting instead of carrying out push-flow iterations until convergence, we perform a fixed number of iterations (i.e. we replace the while with a for loop), and drop nodes whose residual score is below a specified threshold in each iteration. Additionally, we truncate nodes with a very large degree ($\ge 10000$) by randomly sampling their neighbors. The above modifications proved to be just as effective as \citet{andersen2006local}'s method while being significantly faster in terms of runtime.

\subsection{Applicability and Limitations}
\label{sec:limitations}
When using \model~ for your own purposes you should first be aware that this model assumes a homophilic graph, which is mostly, but not always the case. Furthermore, it cannot perform arbitrary message passing schemes like GNNs do, since we essentially compress the message passing into a single step. It therefore has less theoretical expressiveness than GNNs \citep{BoldiLSV06,xu2018powerful}, even if it practically shows the same or better accuracy. However, note that \model~ allows arbitrary realizations of $f_\theta$ and can therefore be used with more complex data and models such as images and CNNs, audio and LSTMs, or text and Transformers.

\end{document}